\documentclass[letterpaper, 10 pt, conference]{ieeeconf}  

\IEEEoverridecommandlockouts                              

\usepackage{graphicx}
\usepackage{color,soul}
\usepackage{dblfloatfix}
\usepackage{multirow}
\usepackage{multicol}
\usepackage{hyperref}
\usepackage{siunitx}
\usepackage{hyperref}
\usepackage{xcolor}
\usepackage{graphicx}
\usepackage{svg}
\usepackage{amsmath}
\usepackage{amsfonts}
\usepackage{amssymb}
\usepackage{array}
\usepackage{url}
\usepackage[noadjust]{cite}

\begin{document}
\title{\LARGE \bf 
Augmented Reality User Interface for Command, Control, and Supervision of Large Multi-Agent Teams}

\author{\protect\parbox{\textwidth}{\protect\centering Frank Regal, Chris Suarez, Fabian Parra, Mitch Pryor}
\thanks{All authors are with the Nuclear and Applied Robotics Group (NRG) at The University of Texas at Austin. Austin, TX, 78712, USA. {\tt \{fregal, mpryor\}@utexas.edu}}
}
\maketitle
\section{OVERVIEW}
Multi-agent human-robot teaming allows for the potential to gather information about various environments more efficiently by exploiting and combining the strengths of humans and robots. In industries like defense, search and rescue, first-response, and others alike, heterogeneous human-robot teams show promise to accelerate data collection and improve team safety by removing humans from unknown and potentially hazardous situations. In recent work, we developed AugRE \cite{regal_augre_2022}, an Augmented Reality (AR) based scalable human-robot teaming framework that enables users to localize and communicate with 50+ autonomous agents. Built as a framework to enable large multi-agent human-robot teaming, studies focused on analyzing the framework's scalability and practicality with limited user interface supervision modalities and no command and control functionalities via the Microsoft HoloLens 2 AR Head Mounted Display (HMD). Through ongoing demonstrations and frequent human-robot teaming tests with AugRE in outdoor urban environments, unexpected events would often transpire, and an autonomous agent would need a manual intervention to redirect or redefine navigation goals to continue to make progress toward the mission goal. It became apparent that users need the ability to supervise, command, and control these agents via the AR headset. This work also exposed the need for users to have the ability to interact with both \textit{close} line-of-sight (LOS) and \textit{remote} non-line-of-sight (NLOS) autonomous agents. Therefore, this work takes the AugRE framework and builds upon it with various AR-based command, control, and supervision modalities. Through our efforts, users are able to interact with large autonomous agent teams without the need to modify the environment \emph{prior} and without requiring users to use typical hardware (\emph{i.e.} joysticks, keyboards, laptops, tablets, etc.) in the field. The demonstrated work shows early indications that combining these AR-HMD-based user interaction modalities for command, control, and supervision will help improve human-robot team collaboration, robustness, and trust. 

\section{BACKGROUND}

AR-HMDs, equipped with onboard compute, vision, depth, and IR sensors, can localize within unstructured environments, track body motion, and offer users digital displays that superimpose virtual information on the physical world. These built-in capabilities make AR-HMDs a practical and efficient solution for connecting and localizing non-expert robotic operators with robotic agents while maintaining portability and situational awareness of the physical world. AR-HMDs additionally provide users with a simple-to-use gesture-based interface, enhancing the practicality of these devices as users can keep their hands free in the field. AR-HMD devices, paired with robotic systems, have proved to enhance user situational awareness of robotic system intentions \cite{ruffaldi2016third}. Additionally, many studies have used AR-HMDs to enable users to command and control unmanned aerial vehicles (UAVs) \cite{hedayati2018improving, huang_augmented_2020}, unmanned ground vehicles (UGVs) \cite{gu_ar_2022, reardon2019augmented}, and manipulators \cite{arboleda2021assisting, regal_augmented_2023}. Most research though has focused on the command, control, and supervision of a \emph{single} robotic platform in controlled environments. 

\begin{figure}[t!]
    \centering
    \vspace{1ex}
    \includegraphics[width=\linewidth]{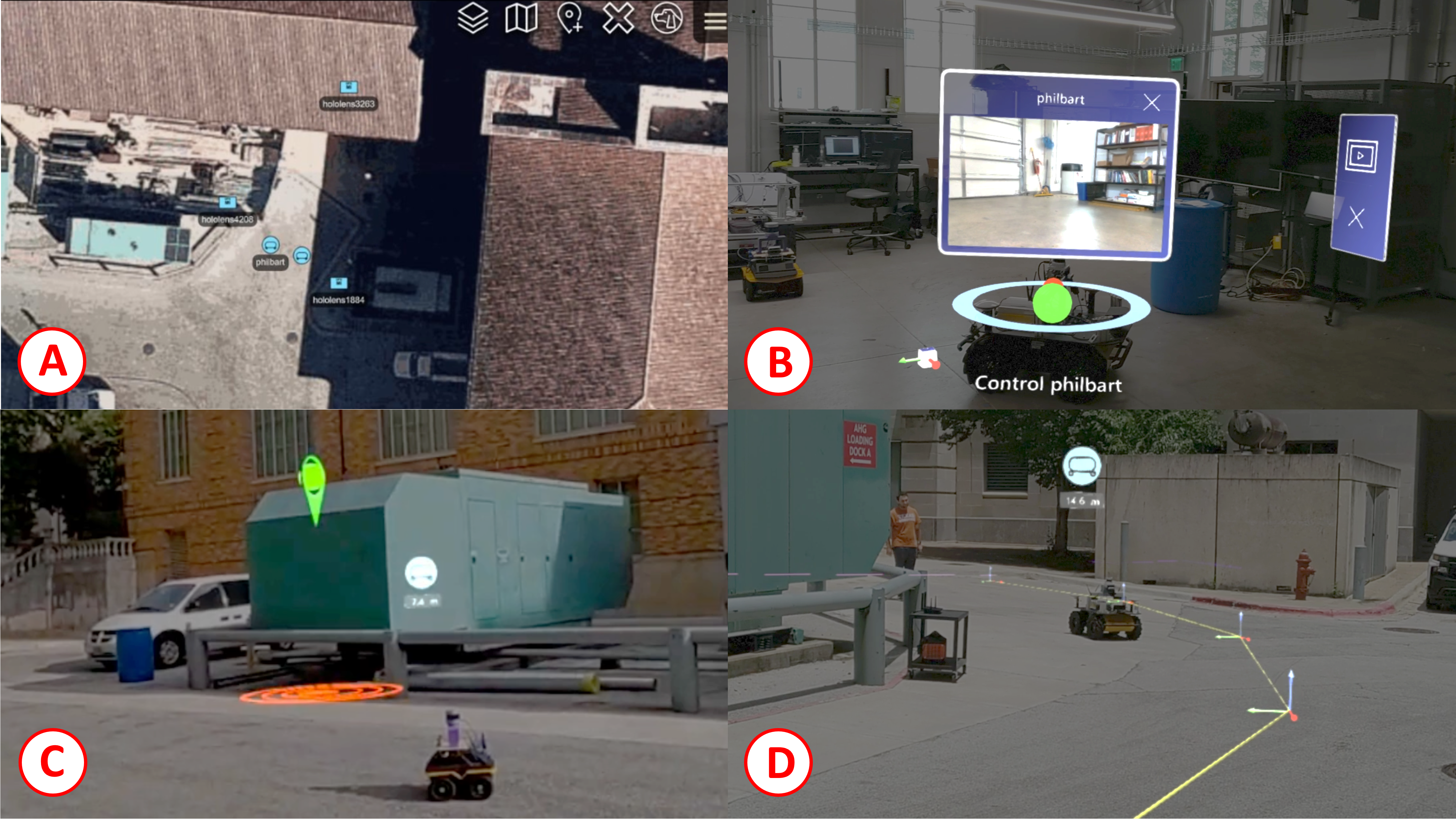}
    \vspace{-3ex}
    \caption{[A] Top-down map view of HoloLens 2 and robotic agent locations. [B] Joystick teleoperation control with FPV robot view for NLOS operation. [C] Autonomous agent navigating to virtual waypoint provided by AR-HMD user. [D] AR-HMD user viewing an agent's navigation path.}
    \label{fig:feature-collage}
    \vspace{-4ex}
\end{figure}

In previous work, we presented Augmented Robot Environment (AugRE) \cite{regal_augre_2022}. This AR-based framework uses Azure Spatial Anchors (ASAs) \cite{arguelles_azure_2022} and Robotfleet \cite{sikand_robofleet_2021} to localize and enable bi-lateral communication between humans and teams of robots. The work is limited in user interface design and only allows users to simply view and interact with LOS autonomous agents and receive notifications from those agents. This work presents an AR-based user interface specifically designed for autonomous agent command, control, and supervision in multi-agent human-robot teams. Demonstrated on hardware\footnote{Demonstration Video: \url{https://utnuclearroboticspublic.github.io/Augmented-Robot-Environment}} in a realistic urban environment, our AR interface enables non-expert users to visualize robot intentions and create navigation plans using novel AR-based waypoint and teleoperation modalities developed to command and control autonomous agents in LOS and NLOS with the user. Specifically, we present the following contributions on hardware, for interaction with LOS and NLOS agents:

\begin{enumerate}
    \item Situational awareness of each agent's relative position \textit{(\ref{sec:ar-label})} and GPS-referenced world positions \textit{(\ref{sec:gps-location})}
    \item Interaction capabilities with \emph{close} LOS and \emph{remote} NLOS agents \textit{(\ref{sec:interactions})}
    \item Live feedback visualizations of autonomous agent video streams along with planned and past trajectories \textit{(\ref{sec:feedback-visuals})}
    \item Natural gesture-based virtual joystick interface for teleoperation \textit{(\ref{sec:teleop})}
    \item Waypoint navigation goal setting and path building interface for semi-autonomous operation \textit{(\ref{sec:waypoint})}
    \item AR user-to-robot leader-follower commanding capability \textit{(\ref{sec:leader-follower})}
\end{enumerate} 

The next section briefly summarizes these capabilities and a link to a video demonstrating these capabilities is given in Section \ref{sec:demonstration}.

\section{USER INTERACTION DETAILS} 
\label{sec:system-overview}
We enable users to command, control, and supervise ROS-based autonomous agents via the AugRE application built for, but not limited to, Microsoft HoloLens 2 AR-HMDs. AugRE uses a) Robofleet \cite{sikand_robofleet_2021} to enable low latency, multi-agent, bilateral communication, b) Microsoft Azure Spatial Anchors (ASA) \cite{arguelles_azure_2022, mcgill_quest_2020} to localize multiple humans and robots in an unknown unstructured environment, and c) Microsoft's Mixed Reality Toolkit (MRTK) \cite{microsoft_mrtk_UX2022} for visualizations to provide situational awareness of all teammates (robots and users). To visualize and interact with agents, it is required for users and autonomous agents first to localize to a common reference point. On team initialization, the first available robot automatically places one Microsoft ASA in the environment. Once created via an RGB sensor, the robot publishes a ROS \texttt{TFMessage} with its pose relative to the ASA. Each additional Microsoft HoloLens 2 user and autonomous agent that join immediately subscribe to the common \emph{transform-tree} being broadcast across a local network to all team members. Each agent queries and matches the ASA saved spatial information, referenced as a base frame in the \emph{transform-tree}. As spatial information is matched, agents add and continuously publish their relative location with respect to the closest ASA via a \texttt{TFMessage} in the team \emph{transform-tree} (see \cite{regal_augre_2022} for more detail). After each agent is localized and connected to a common network, users can begin to visualize and interact with autonomous agents via the command and control interface modalities developed here. The following sections discuss these AR-HMD-based user interface modalities in more detail.

\subsection{Situational Awareness}
\label{sec:sit-awareness}
In large human-robot teams, it is desirable to visualize world agent locations and have the ability to interact with those teammates from anywhere in the environment. AR-labels are used here to visualize agent-relative and GPS-referenced world positions. Users can improve their situational awareness of \emph{close} LOS and \emph{remote} NLOS autonomous agents by visualizing live video sensor streams and viewing planned and previously traveled navigation paths. Details of each modality are explained below.

\begin{figure}[t!]
    \centering
    \vspace{1ex}
    \includegraphics[width=\linewidth]{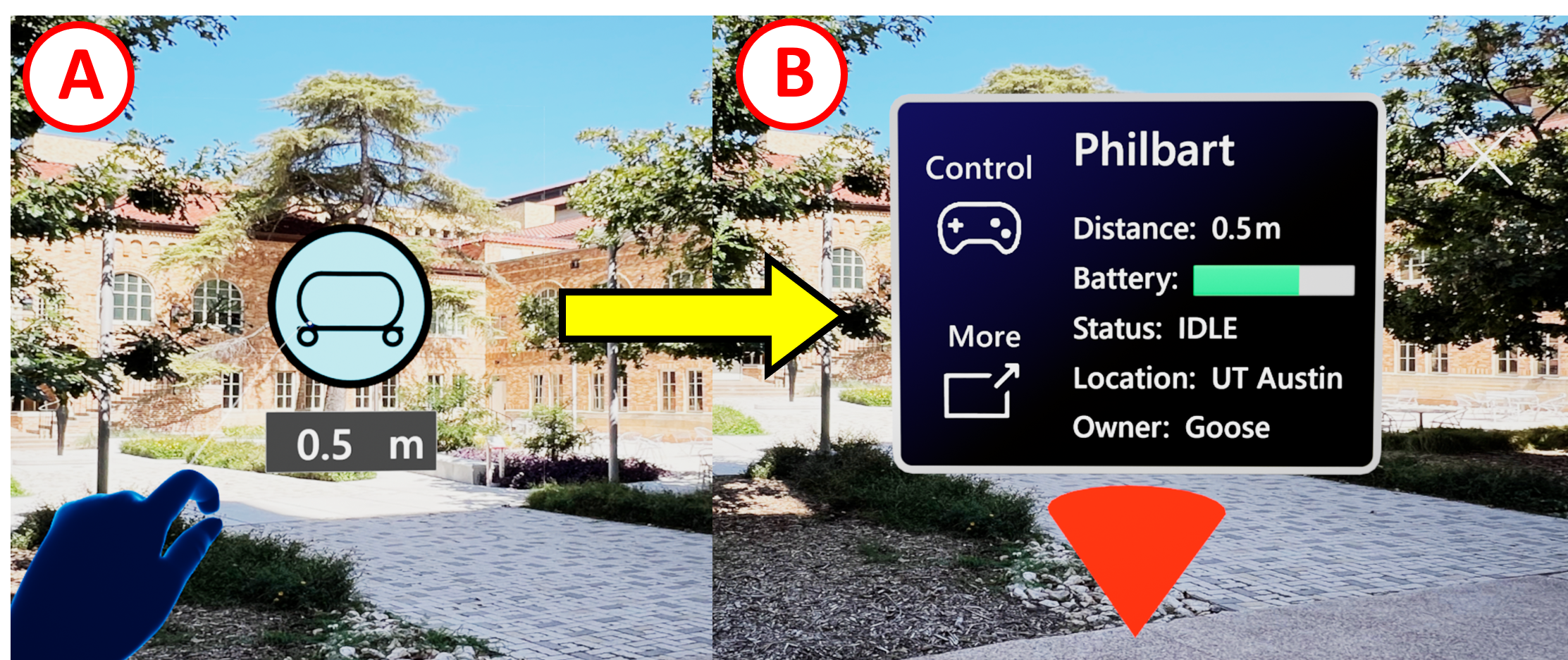}
    \vspace{-3ex}
    \caption{[A] AR-label used to visualize agent type and location in the environment. Blue NATO symbol represents an unmanned ground vehicle (UGV). [B] AR-label, expanded with click, providing additional information and clickable ``Control" and ``More" buttons to access more control, command, and supervision functionalities.}
    \label{fig:ar-label}
    \vspace{-4ex}
\end{figure}

\subsubsection{World positioned AR-labels}
\label{sec:ar-label}
AR-labels used to visualize agent locations are interactable holographic buttons. These labels are visualized for each connected agent through the AR-HMD based on the relative ASA pose published as a \texttt{TFMessage} in the team \emph{transform-tree}. When clicked using a gesture-based air tap, the AR labels provide the user additional teammate information and command and control functionalities (Fig. \ref{fig:ar-label}). For autonomous agents, the information on the expanded menu includes battery life, control mode status (\emph{i.e.} idle, teleoperation, autonomous, etc.), owner, and the normal Euclidean distance relative to the user. Additional holographic buttons exist on this expanded information menu for a user to view a robot's live vision feed, teleoperate the robot, and command the robot with navigation waypoint goals (discussed further below). For other AR-HMD users, only the user's name is currently displayed when clicks are made on the base AR-label. Information that fills AR-label information menu fields is from a custom \texttt{AgentStatus} message that each team member publishes when connected.

\subsubsection{Close \& remote interactions}
\label{sec:interactions}
Clickable interactions with the holographic AR-label buttons are not accessible using built-in HoloLens 2 mixed-reality interactions beyond \SI{5}{\m}. This work assumes that teammates could be located at relative distances beyond \SI{50}{\m}. Therefore, we developed a far-distance interaction mode for the clickable AR-labels, automatically enabled when agents are located outside a threshold. AR-labels begin to scale down for agents with a normal Euclidean distance of \SI{3}{\m} or more to provide the illusion the agent is traveling far away. At \SI{4.5}{\m} and beyond, the AR-label scale-down stops and \emph{detaches} from the agent's world position. At this point, an agent's AR-label is physically rendered on the AR-HMD display at \SI{4.5}{\m}, scaled down by a factor of \SI{2}{}. The label is oriented and aligned with the straight-line vector that exists between the two teammates, giving each user the effect that the label is located on top of the far-distance agents, while still visible and intractable because it is within the bounds of interactability. A resulting capability prevents AR-labels from colliding or becoming occluded by buildings or other objects in the environment that may be blocking LOS between two teammates. This functionality enables users to have the ability to interact and view information about autonomous agents or other users who are located NLOS behind buildings or large objects in the environment. 

\subsubsection{GPS-referenced world positions}
\label{sec:gps-location} 
We also provide the capability to operate with GPS-referenced coordinates.  At least one ASA in the scene, a priori, is assigned a GPS coordinate with proper heading. Agents that localize against any ASA - not necessarily the one with assigned coordinates - can then extrapolate their respective GPS location.  We demonstrate this utilizing the publicly available Android Team Awareness Kit (ATAK) (Fig. \ref{fig:feature-collage}-A). No agent in our demonstration uses a GPS sensor to report coordinates. Instead, they rely on this method, which is generally more precise in an urban setting and operates in entirely GPS-denied areas. This functionality also works in the opposite direction. If an agent cannot localize in reference to an ASA and is equipped with a GPS sensor, its relative location and other awareness data can be visualized and interacted with in our AR-HMD setup as with any other team member.

\subsubsection{Feedback visualizations}
\label{sec:feedback-visuals}

\begin{figure}[t!]
    \centering
    \vspace{1ex}
    \includegraphics[width=0.8\linewidth]{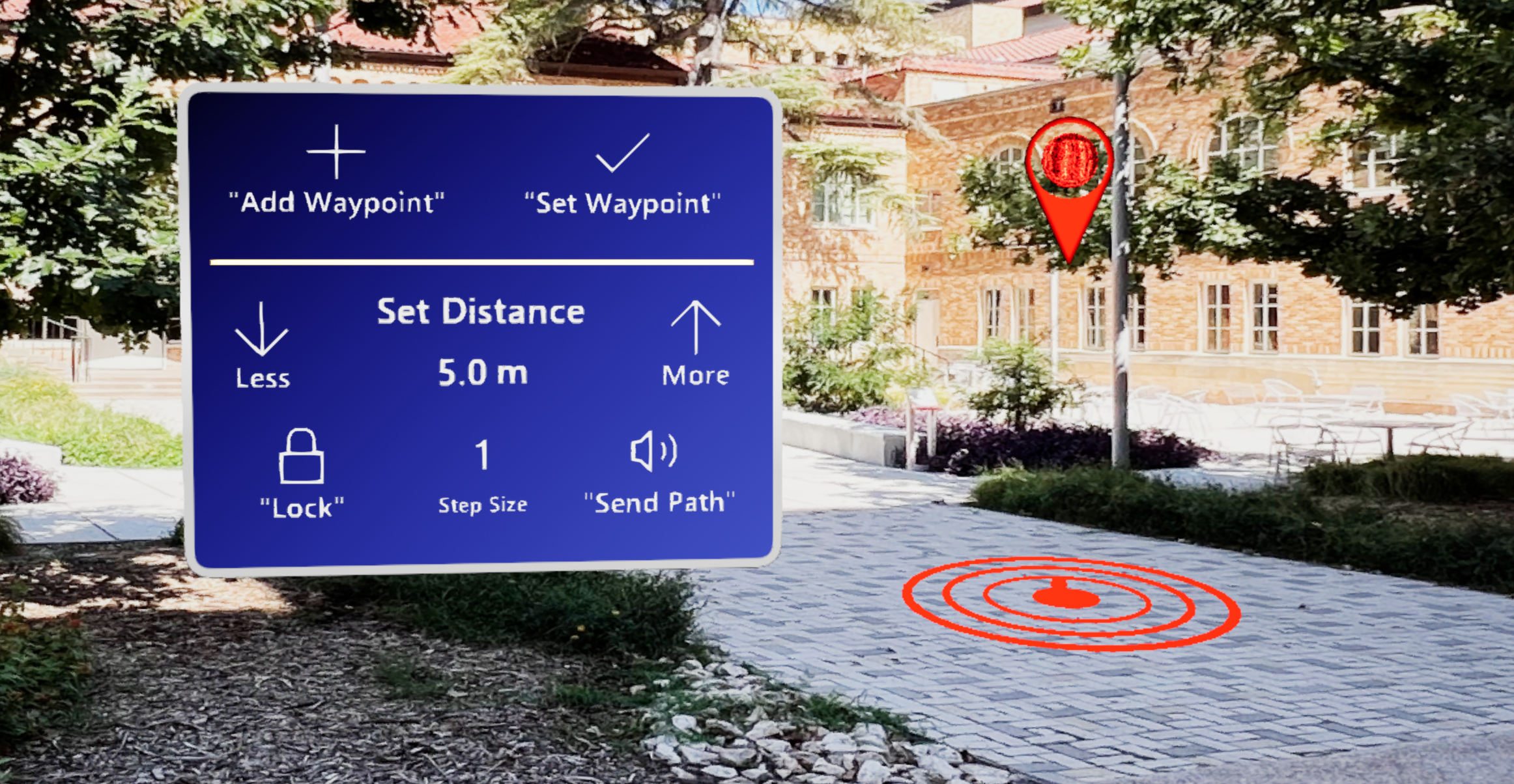}
    \vspace{-1ex}
    \caption{Holographic waypoint options menu and red holographic marker presented to users to physically define a waypoint navigation goal for autonomous agents.}
    \label{fig:waypoint-menu}
    \vspace{-4ex}
\end{figure}

To further enhance users' situational awareness relative to remote agents that may be NLOS, we provide the capability to open a live video stream on a holographic panel on the user's AR-HMD (Fig. \ref{fig:feature-collage}-B). Users can view and cycle through each RGB sensor video feed available from an autonomous agent if equipped. AR-HMD users listen to ROS-based \texttt{sensor\_msgs/CompressedImage} messages for each autonomous agent to visualize these streams. When users open the expanded AR-label (Fig. \ref{fig:ar-label}-B), users can click on a ``Live View" button to open a video stream holographic panel with a cycle button. With decoding, decompressing, and rendering video stream content on a low-compute AR-HMD, video streams are only rendered at whatever frame is available every \SI{10}{\Hz}.
In addition to visualizing video streams, users can also visualize an agent's projected and past trajectory path when autonomously navigating to provide users with advanced awareness of where the autonomous agent is about to travel. Each agent enables this by publishing a ROS \texttt{nav\_msgs/Path} message. On the AR-HMD, the device linearly interpolates the information and renders paths aligned to the ground of the environment (Fig. \ref{fig:feature-collage}-D).

\subsection{Robot Control}
\label{sec:robot-control}

\subsubsection{Teleoperation}
\label{sec:teleop}
Users can manually intervene with an autonomous agent's current plan by clicking the ``Control" button on the expanded AR-label (Fig. \ref{fig:ar-label}-B). Once clicked, users are presented with a holographic joystick aligned relative to the user's current position and orientation (see Fig. \ref{fig:feature-collage}-B). Using the built-in HoloLens 2 pinch-and-hold hand gesture, users can manipulate the green sphere to control the agent. For non-holonomic vehicles, we applied a constraint on movement so the user can translate the sphere in the $x$- and $y$-direction, and rotate it about the $z$-axis (yaw). Linear and angular velocities are calculated using the holographic sphere's initial and current centroid poses over the delta time of each rendered frame. These velocities fill a ROS \texttt{geometry\_msgs/Twist} message and are sent to the respective agent. To teleoperate \emph{remote} NLOS agents with the joystick, an additional ``Live View" button is added to the user's environment when the joystick modality is opened (Fig. \ref{fig:feature-collage}-B). When clicked, the live video feed (discussed in Section \ref{sec:feedback-visuals}) is rendered in the user's view. This allows users to teleoperate agents with a FPV. When a user begins to manipulate the virtual joystick, the AR-label for that autonomous agent in each user's view turns red, providing instant control awareness for teammates. In the expanded AR-label version, the ``status" field changes to ``teleoperation" and the ``owner" field shows the name of who is commanding the agent.


\subsection{Robot Commanding}
\label{sec:robot-commanding}

\subsubsection{Waypoint Navigation}
\label{sec:waypoint}
\begin{figure}[t!]
    \centering
    \vspace{1ex}
    \includegraphics[width=0.8\linewidth]{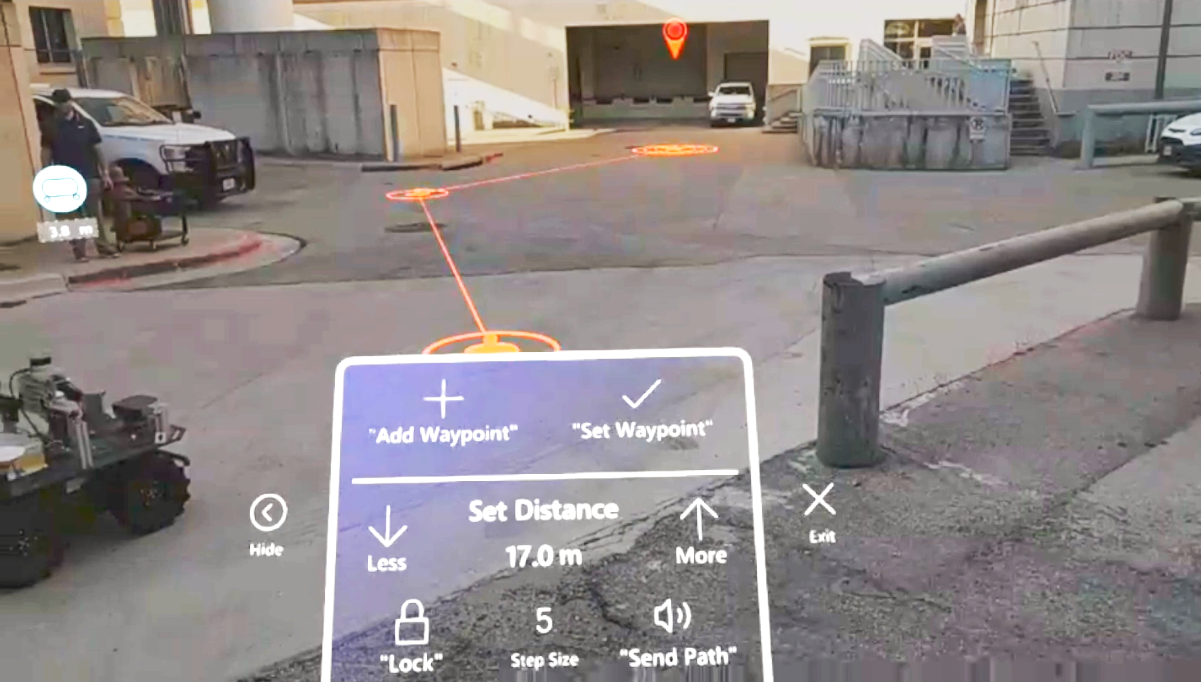}
    \vspace{-1ex}
    \caption{User creating a waypoint navigation path using the waypoint feature menu. Three waypoints were set to command the Clearpath husky pictured in the bottom left to travel to the end of the alleyway.}
    \label{fig:waypoint-path}
    \vspace{-4ex}
\end{figure}

With this interface, users can create a single navigation waypoint goal (Fig. \ref{fig:feature-collage}-C and Fig. \ref{fig:waypoint-menu}) or create a set of waypoints (Fig. \ref{fig:waypoint-path}) to provide autonomous agents with a path to follow. Users can open the navigation path feature by clicking through the expanded AR-label. On open, users are presented with a red holographic point marker, a red target grounded to the physical environment, and an options menu (see Fig. \ref{fig:waypoint-menu}). The red holographic point marker and target are aligned vertically with each other. The two are placed \SI{5}{\m} away from the user along a straight-line vector extruded from the centroid of the AR-HMD. The marker stays aligned and centered with the user's head as the user rotates about the $z$-axis (yaw). Through a ``Lock/Unlock" voice command, users can lock and unlock the marker's physical position in the world. With clicks through the associated options menu, the user can increase or decrease the distance the marker is located relative to oneself. This allows users to place waypoints at long distances (\emph{i.e.} \SI{100}{\m} or more), which is desirable for users who are exploring large open environments. To build a path, users can use the ``Set Waypoint" or ``Add Waypoint" voice commands or click through the associated buttons on the options menu. On waypoint set or add, the previous waypoint in the scene locks in its physical location and allows users to set a new one. As the path is being built, a holographic line is visualized between each waypoint, giving users awareness of the sequence of waypoints. When a user is satisfied with the goal or waypoint path created, users can voice "Send Goal" or "Send Path" or click the associated buttons. Once sent, an array of poses for each holographic marker are transformed into the closest ASA reference frame and sent to the respective robot via a ROS \texttt{nav\_msgs/Path} message. Additionally, the markers located in the environment change colors, and a confirmation sound plays to notify and confirm with the user the command sent. 

\subsubsection{AR user-to-robot leader-follower control}
\label{sec:leader-follower}
In tense situations when users need to command an autonomous agent rapidly, it may be undesirable to teleoperate via the holographic joystick or build a waypoint path via the waypoint navigation feature discussed above. Therefore, we allow users to command a single or group of autonomous agents to follow a user as they navigate through the environment. By appropriately selecting which robot to command, via clicks on the expanded AR-label, users can voice "follow me," and the autonomous agent will begin to follow the user at a predefined location, \SI{1}{\m} directly behind the user. This pose is updated and sent to the robot every \SI{10}{\Hz}. Once transformed to the closest ASA reference frame, a ROS \texttt{geometry\_msgs/PoseStamped} is continuously sent to the autonomous agent, and the agent begins to navigate to that location, updating its goal on every new message received. For the user, a holographic marker is placed in the environment at the current goal location, and an audible confirmation noise to confirm to the user the ``follow me" command is recognized. This provides users with direct feedback to ensure they know where the robotic agent travels. To stop the follow command, users can provide a ``Stop" voice command to stop the autonomous agent from following the user. 

\section{TEST IMPLEMENTATION} 
\label{sec:applications}
The AR-HMD-based user interface is demonstrated on hardware in an outdoor urban, unknown environment. Three AR-HMD users and two autonomous UGV's were tasked to explore a large area (roughly \SI{200}{\m} x \SI{400}{\m} area) and find a blue barrel that was hidden before the demonstration. One Clearpath Jackal and one Clearpath Husky equipped with LiDAR and RGB-D sensors for localization and object recognition were used. Each agent within the team was assigned a task. One AR-HMD user was responsible for supervising the entire team from the ground station base location, located just inside a garage. The other two fielded AR-HMD users were responsible for commanding and controlling each autonomous agent in the field. Both autonomous UGVs were responsible for traveling to locations in the environment their fielded teammates desired and continuously searched for blue barrels. At the start of the exploration task, the fielded users each commanded a UGV to travel to a far-distant waypoint goal or path. The users collectively set distant goals for the UGVs to access areas of the environment not initially visible to all agents. While the UGVs moved towards these waypoints, the autonomous agents continuously scanned the surroundings for a blue barrel. Once a UGV spotted the barrel, all team members received a notification containing an image of the object and a holographic marker pinpointing its location from each user's perspective. From the base station, the team supervisor utilized NLOS interactions to activate the teleoperation mode of the autonomous agent who detected the barrel. Using an FPV video stream from the UGV, the supervisor controlled the robot approximately \SI{200}{\m} away to inspect the barrel more closely. Following the detailed inspection, each team member commanded the nearest robot to return to the base using the ``follow me" command feature. Each agent in the team reconvened back at base to complete the mission.

\section{CONCLUSION} 
\label{sec:results}
We present an AR-HMD-based user interface designed for commanding, controlling, and supervising large multi-agent teams. Built on AugRE \cite{regal_augmented_2023}, our interface facilitates users in visualizing real-world locations of all agents through world-aligned AR-labels. Regardless of whether agents are within the users' LOS or NLOS view, these clickable AR-labels provide access to various command, control, and supervision options. Our user interface offers a teleoperation mode to control agents, allowing users to teleoperate an agent using pinch-and-hold gestures on a holographic mesh. Additionally, by enabling a live view video stream, users can gain an FPV of the robot displayed on a holographic panel in the AR-HMD. Developed waypoint goals and path-building functionalities enable users to set waypoints for autonomous agents to follow. These waypoints can be placed at considerable distances, allowing agents to access and survey unexplored parts of the environment for better observation. Finally, we present a leader-follower modality, enabling users to command a robot to follow them as they navigate to different locations in the environment. This comprehensive AR-HMD interface empowers users with versatile tools to manage and coordinate multi-agent teams efficiently. Future work will perform studies to determine the interface's ability to improve team collaboration, robustness, and trust with users.

\section{DEMONSTRATION}
\label{sec:demonstration}
\centering \url{https://utnuclearroboticspublic.github.io/Augmented-Robot-Environment/}


\break
\typeout{}
\bibliographystyle{IEEEtran}
\bibliography{IEEEabrv,bibliography}



\end{document}